\documentclass[10pt,twocolumn,letterpaper]{article}

\usepackage[pagenumbers]{cvpr}
\makeatletter
\renewcommand{\LenToUnit}[1]{\strip@pt\dimexpr#1\relax}
\makeatother
\usepackage{times}
\usepackage{fix-cm}
\def\confName{CVPR}
\def\confYear{2027}
\usepackage{xr-hyper}
\usepackage[pagebackref]{hyperref}
\usepackage{url}
\usepackage{graphicx}
\usepackage{booktabs}
\usepackage{needspace}
\usepackage{multirow}
\usepackage{amsmath,amssymb,mathtools,amsthm}
\usepackage{xcolor}
\usepackage{array,colortbl}
\usepackage{microtype}
\usepackage{enumitem}
\usepackage{adjustbox}
\usepackage{caption}
\usepackage{float}
\usepackage{flafter}
\usepackage{placeins}
\usepackage{wrapfig}
\usepackage{tabularx}
\usepackage{algorithm}
\usepackage{algpseudocode}

\floatstyle{ruled}
\restylefloat{algorithm}

\hypersetup{hidelinks,
  pdftitle={MapSelect: Mapping Preservation and Adapter Selection for Continual Multimodal Instruction Tuning},
  pdfauthor={Qian Feng, Xiyang Du, Zhikai Chen, Chao Zhang, Peng Zhang, Hui Qian},
  pdfsubject={Research preprint},
  pdfkeywords={continual learning, multimodal instruction tuning, low-rank adaptation}}

\definecolor{diagblue}{HTML}{326DB5}
\definecolor{diagorange}{HTML}{DC8440}
\definecolor{diagpurple}{HTML}{9274B5}
\definecolor{diaggreen}{HTML}{27947D}

\newcommand{\method}{{\upshape\scshape MapSelect}}

\definecolor{tabshade}{HTML}{D5E6F8}
\definecolor{tabhead}{HTML}{EEF1F5}
\definecolor{tabstripe}{HTML}{F3F7FC}
\definecolor{tabgroup}{HTML}{E5EFF9}
\definecolor{tabteal}{HTML}{E1F2EC}
\definecolor{tabink}{HTML}{285F9B}

\author{Qian Feng \quad Xiyang Du \quad Zhikai Chen\\Chao Zhang \quad Peng Zhang \quad Hui Qian}

\title{MapSelect: Mapping Preservation and Adapter Selection\\for Continual Multimodal Instruction Tuning}
\makeatletter
\g@addto@macro\@maketitle{%
  \begin{minipage}{\textwidth}
  \centering
  \includegraphics[width=\textwidth]{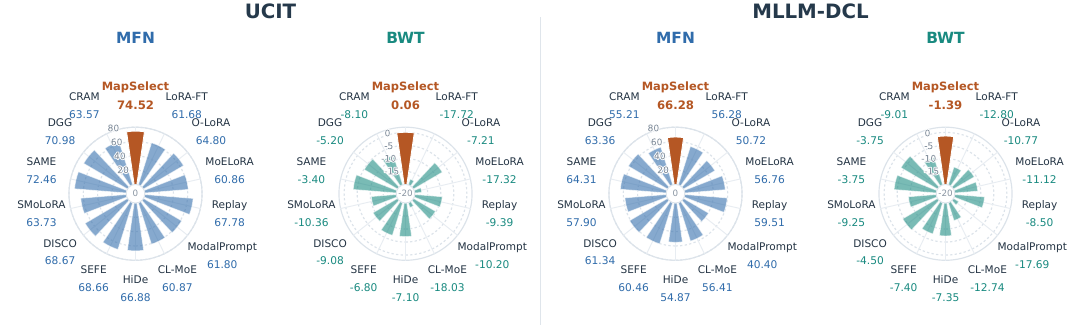}
  \par\vspace{-1mm}
  {\captionsetup{type=figure,justification=raggedright,singlelinecheck=false,labelfont={bf,color=black},textfont={color=black}}
  \captionof{figure}{Final performance and backward transfer of 14 methods on UCIT (left) and MLLM-DCL (right). Bars and outer numbers show the MFN and BWT scores reported in Tables~\ref{tab:ucit} and~\ref{tab:dcl}. \method{} is highlighted in orange. Higher scores extend farther from the center, with matched radial scales across benchmarks for each metric.}
  \label{fig:benchmark-radar}}
  \end{minipage}\par\vspace{1mm}
}
\makeatother
\begin{document}
\maketitle
\begin{abstract}
Continual multimodal instruction tuning enables multimodal large language
models to acquire new capabilities, but sequential adaptation often
degrades performance on earlier tasks. Existing methods retain task-wise
language adapters, whose effective use also depends on the visual mapping
and adapter selection. We investigate these factors through controlled
projector and expert interventions and show that both can recover
forgotten answers. Based on these observations, we propose \method{},
which combines mapping preservation with adapter selection.
History-Guided Mapping Adaptation (HMA) retains task-specific projector
adapters and guides new updates with historical mapping directions.
Selection-Guided Adapter Reuse (SAR) incorporates related historical
language adapters during learning and maintains prompt-conditioned
selection throughout each answer. Pretrained Subspace Guidance (PSG)
further constrains language updates with pretrained weight directions.
Experiments on UCIT and MLLM-DCL achieve Mean Final Accuracy of 74.52
and 66.28, respectively. Ablations and controlled comparisons demonstrate
the benefits of visual adaptation, historical adapter reuse, and
pretrained guidance for continual learning.
\end{abstract}
\section{Introduction}
\label{sec:introduction}

Visual instruction tuning~\citep{llava2023,llava15_2024} enables
multimodal large language models (MLLMs) to follow natural-language
instructions for image recognition, visual question answering, and
image captioning. Practical applications require these models to
acquire new capabilities as tasks arrive sequentially, motivating
continual multimodal instruction
tuning~\citep{coin2024,mcitlib2025,mllmcl2025}. However, adapting to new
tasks can degrade performance on earlier tasks, leading to
\emph{catastrophic forgetting}~\citep{mccloskey1989catastrophic,ratcliff1990connectionist}.

\begin{figure*}[t]
\centering
\includegraphics[width=\textwidth]{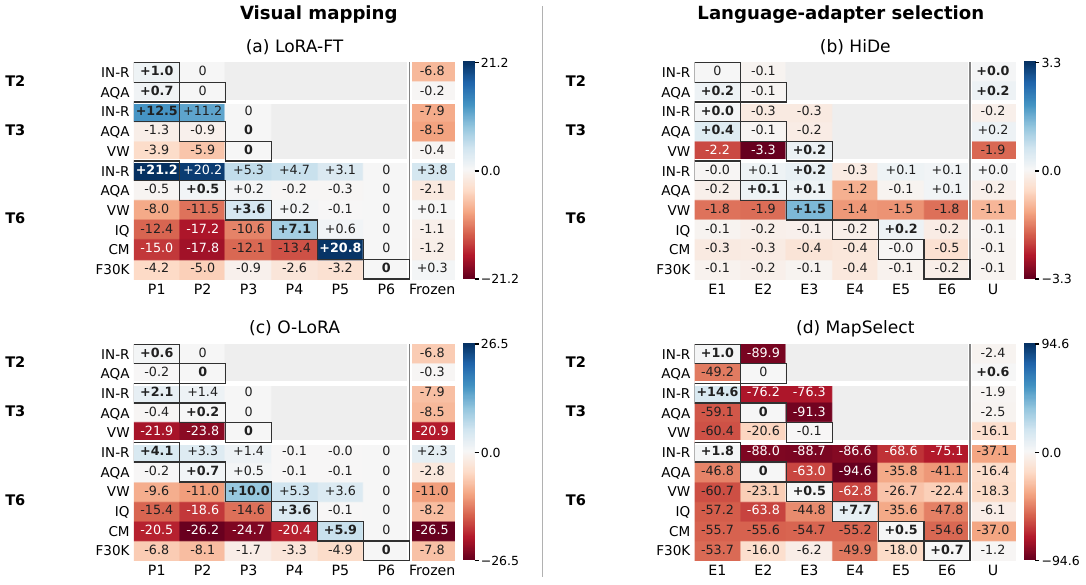}
\caption{\textbf{Visual mappings and language-adapter selection affect retained performance.}
UCIT order is ImageNet-R (IN-R) $\rightarrow$ ArxivQA (AQA) $\rightarrow$
VizWiz (VW) $\rightarrow$ IconQA (IQ) $\rightarrow$ CLEVR-Math (CM)
$\rightarrow$ Flickr30k (F30K). Blocks show performance after learning tasks 2, 3, and 6 (T2, T3, and T6).
Left, $P_j$ substitutes the projector saved after task $j$ with
language-model parameters fixed. Frozen reports frozen-minus-trainable
scores from a separate paired-training comparison. Right, $E_j$ forces expert $j$ and the separate U column uniformly
mixes language experts, with the native visual mapping fixed.
$P_j$, U, and $E_j$ entries are changes from the native configuration.
Boxes mark task correspondence, not a guaranteed optimum. Forced experts
are diagnostic interventions; deployed selection uses input features. Bold marks the best
fixed-checkpoint score per row, including the native reference, using
unrounded values. Gray marks future adapters. Each panel has its own zero-centered color scale.
QA uses percentage points and VW/F30K use CaptionAvg points.}
\label{fig:mapping}\label{fig:selection}
\end{figure*}

Recent approaches use low-rank adaptation (LoRA) to learn
task-specific adapters in frozen MLLMs and retain them for subsequent
use~\citep{hu2022lora,hide2025,smolora2025,same2026}. Retaining these
parameters preserves specialized transformations, but their contribution
to a prediction also depends on the visual representations they receive
and the adapters selected for the input. Both can change as the model
learns new tasks. This raises a question for continual multimodal
learning. \emph{How can retained adaptations remain effective as the
visual mapping and the set of available adapters evolve?}

To answer this question, we investigate visual mappings and
language-adapter selection through controlled interventions on UCIT
and MLLM-DCL. As illustrated in the left panels of
Figure~\ref{fig:mapping}, restoring a historical projector while
keeping language-model parameters fixed can recover performance on
earlier tasks. Different tasks favor different historical projectors,
indicating that the visual mapping learned for a new task can become
less suitable for an earlier one. The Frozen columns further compare
models trained with frozen and trainable projectors. These observations
motivate \textbf{mapping preservation}, which maintains task-specific
visual mappings throughout continual learning.

We then examine whether the retained language adapters are effectively
used. With the visual mapping and model parameters fixed, we substitute
individual historical experts for the original adapter selection.
As illustrated in the right panels of Figure~\ref{fig:selection},
selecting a task's retained adapter can recover earlier-task
performance. Uniform mixing provides a comparison that uses all
available experts. The recovery under individual experts shows that
useful adaptations remain available after subsequent learning and that
\textbf{adapter selection} affects access to these capabilities.

These findings motivate \method{}, which jointly preserves visual
mappings and improves the use of retained language adapters.
History-Guided Mapping Adaptation (HMA) retains task-specific projector
adapters and uses directions learned by earlier adapters to guide new
mapping updates. Selection-Guided Adapter Reuse (SAR) incorporates a
related historical language adapter during new-task learning and
maintains prompt-conditioned adapter selection throughout each answer.
Pretrained Subspace Guidance (PSG) further constrains language updates
with pretrained weight directions. HMA and PSG apply a common soft
gradient constraint to historical visual adaptations and pretrained
language knowledge, respectively. The main contributions are summarized
as follows.
\needspace{5\baselineskip}
\begin{itemize}[leftmargin=*,itemsep=1pt,topsep=2pt]
\item We investigate mapping preservation and adapter selection in
continual multimodal instruction tuning. Projector and expert
interventions show that both factors affect the recovery of forgotten
answers.
\item We propose \method{}, combining history-guided mapping adaptation,
selection-guided adapter reuse, and pretrained subspace guidance to
preserve and use task-specific adaptations.
\item Experiments on UCIT and MLLM-DCL demonstrate improved final
performance and retention. Ablations and controlled comparisons examine
the effects of visual adaptation, historical reuse, and pretrained
guidance.
\end{itemize}

\section{Related Work}
\label{sec:related}
\noindent\textbf{Continual multimodal instruction tuning.}
CoIN~\citep{coin2024}, MCITlib~\citep{mcitlib2025}, and
MLLM-DCL~\citep{mllmcl2025} provide benchmarks for sequential learning
across multimodal tasks and domains. Existing methods reduce
interference through orthogonal low-rank updates, task-specific experts,
and modality-aware prompts. O-LoRA~\citep{olora2023} separates update
subspaces, HiDe-LLaVA (HiDe)~\citep{hide2025} decouples task adaptations, and
CL-MoE~\citep{clmoe2025} combines experts for continual visual question
answering. ModalPrompt~\citep{modalprompt2025} uses dual-modality
guidance, while SMoLoRA~\citep{smolora2025} separates visual
understanding from instruction following. DGG~\citep{dgg2025}
approximates old-task gradients to guide new-task learning. Our work
examines how changes in visual mappings and language-adapter selection
affect the continued use of retained adaptations.

\medskip
\noindent\textbf{Visual adaptation.}
LLaVA~\citep{llava2023,llava15_2024} and
InternVL~\citep{chen2024internvl} connect visual encoders to language
models through learned projectors. MVP~\citep{jin2025mvp} uses
instruction-conditioned projectors, and PMA~\citep{pma2026} expands
projector experts to preserve multimodal alignment. These studies
establish the importance of adapting the visual interface. HMA retains
task-wise projector adapters and constrains new updates using historical
mapping directions. The projector interventions provide an empirical
basis for preserving these alternative mappings.

\medskip
\noindent\textbf{Expert selection and reuse.}
SAME~\citep{same2026} addresses router and expert drift through
stabilized selection and constrained expert updates.
CRAM~\citep{tang2026cram} combines centroid-guided routing with adaptive
expert allocation and orthogonality constraints. Concurrent
CART~\citep{cart2026} studies compatibility-aware selection of projector
and language experts. These methods highlight the relationship between
expert learning and retrieval. SAR uses related historical adapters
during training and prompt-conditioned selection during generation.
Together with HMA, it addresses the visual representations supplied to
language adapters and the selection of those adapters.

\section{Mapping Preservation and Adapter Selection}
\label{sec:diagnosis}
We study how visual mappings and language-adapter selection affect
earlier-task performance after continual learning. Projector
interventions keep language-model parameters fixed, and expert
interventions additionally keep the visual mapping fixed. All
interventions use the same evaluation inputs, decoding, and scoring.

\subsection{Mapping Preservation}
\label{sec:mapping-diagnosis}
As illustrated in the left panels of Figure~\ref{fig:mapping},
restoring historical projectors recovers performance for many earlier tasks in
both LoRA-FT and O-LoRA. For O-LoRA at T3, restoring the first-task
projector recovers 96.1\% of the forgotten ImageNet-R answers.
Task correspondence identifies a candidate mapping; its usefulness depends
on compatibility with the task and the retained language parameters.
The language-model parameters are identical across these substitutions,
so this recovery directly demonstrates the effect of the visual mapping
on retained performance.

The preferred mapping also depends on the task. At the end of UCIT,
earlier LoRA-FT projectors improve ImageNet-R, while the newly learned
Flickr30k task favors the current projector. O-LoRA exhibits a similar
task-dependent pattern. This motivates preserving task-wise visual
mappings so that earlier tasks can continue to use compatible
representations as the model learns new tasks.

The Frozen columns evaluate the alternative of keeping the projector
fixed throughout training. Each measured entry compares separately trained
frozen- and trainable-projector models. This tests the learning tradeoff
of a fixed mapping, whereas projector substitution tests recovery within
one language-model checkpoint. Table~\ref{tab:frozen-projector}
summarizes the final performance of the paired training models.

\subsection{Adapter Selection}
\label{sec:selection-diagnosis}
We next examine whether retained language adapters can still support
earlier tasks when selected appropriately. As illustrated in the right
panels of Figure~\ref{fig:selection}, substituting individual historical
experts for the original selection can improve earlier-task scores.
The effect is more pronounced in \method{} than in HiDe, whose
expert substitutions generally produce smaller changes. Uniform mixing
uses the same expert bank and tests input-independent selection; it
need not be inferior for every task.

On ImageNet-R at T3, selecting its retained adapter in \method{}
increases accuracy from 76.30\% to 90.87\%. This improvement requires
no parameter updates, showing that the earlier adaptation remains
useful after subsequent learning. It also exposes a gap between available
adapter capability and the native selector. Forced selection diagnoses
that gap; inference uses input features without task labels. The projector and expert
interventions together motivate preserving compatible visual mappings
and improving access to historical language adaptations. Further results about projector restoration and adapter selection are provided in Appendices~\ref*{app-main:a_mapping} and~\ref*{app-main:b_selection}.

\section{Mapping Preservation and Adapter Selection with \method}
\label{sec:method}

\begin{figure*}[t]
\centering
\includegraphics[width=\textwidth]{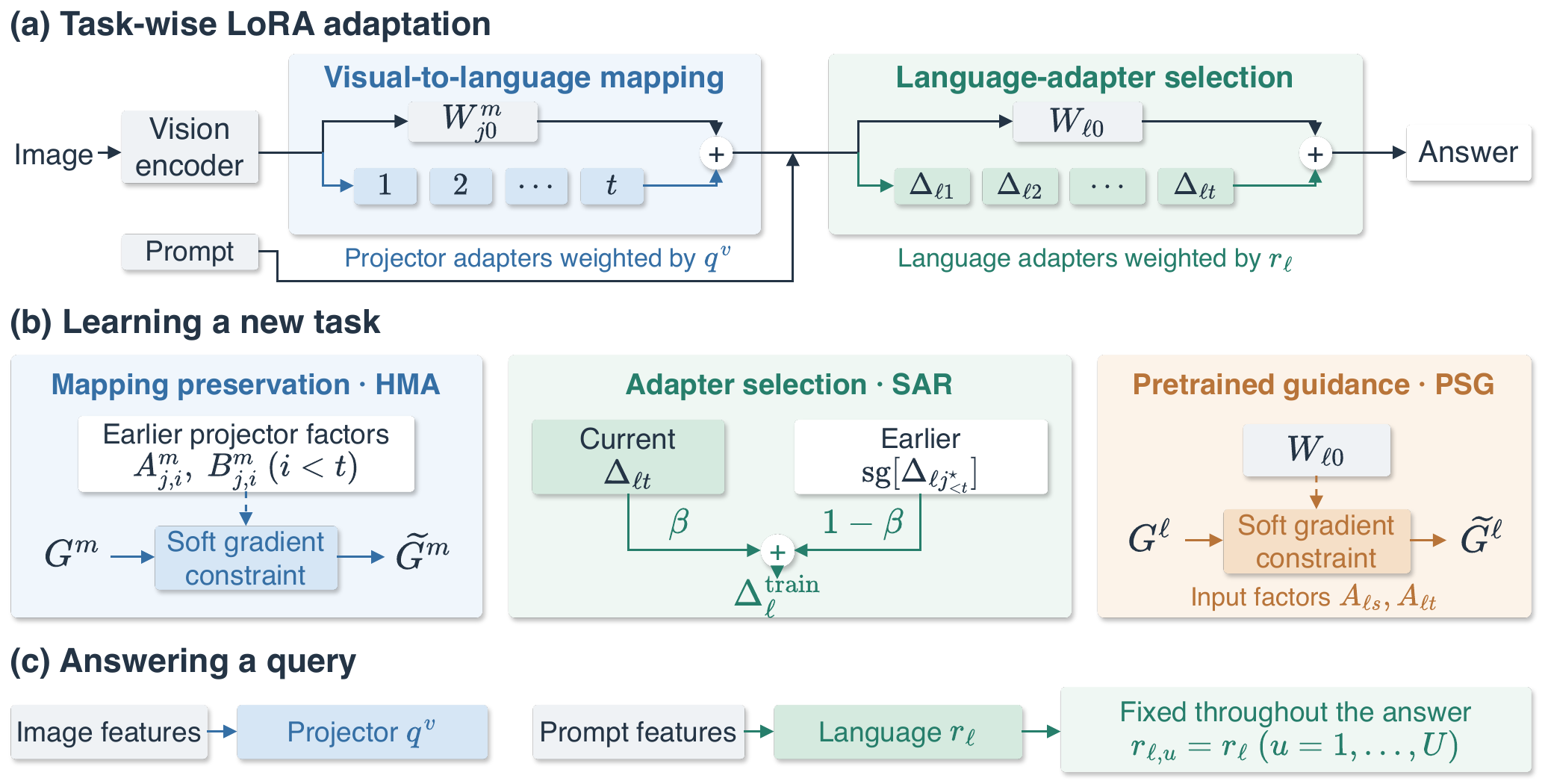}
\caption{\textbf{Overview of \method.}
(a) Task-wise adapters adapt the projector and language model while
pretrained weights remain frozen. (b) HMA constrains projector gradients
using earlier factors. SAR combines current and earlier language
adapters. PSG constrains language input-factor gradients using pretrained
weights. $G^m$ and $G^\ell$ denote gradients, tildes denote filtered
gradients, and $\operatorname{sg}$ stops gradient flow. (c) Image and
prompt features select the respective adapters without task labels.
Language choices are made separately at each layer and stay fixed
throughout the answer.}
\label{fig:diagmoe-overview}
\end{figure*}

\method{} combines mapping preservation and adapter selection through
HMA, SAR, and PSG, as illustrated in
Figure~\ref{fig:diagmoe-overview}. We first introduce the task-wise
adapters, then describe how the three components guide their learning
and use. Algorithm~\ref{alg:mapselect} summarizes training and inference.

\subsection{Task-Wise LoRA Adapters}
We adapt a pretrained MLLM with task-wise LoRA adapters in the visual
projector and language model. The pretrained weights remain frozen.
At task $t$, the model learns new task-specific factors while retaining
the completed task-specific factors for subsequent selection and reuse.

For projector transformation $j$, let $W^m_{j0}$ be the pretrained
weight and $(A^m_{j,i},B^m_{j,i})$ be the adapter factors for task $i$.
Given visual features $z$, the adapted transformation is
\begin{equation}
 m_j(z;q^v)=W^m_{j0}z+
 \eta_m\sum_{i=1}^{t}q_i^v B^m_{j,i}A^m_{j,i}z,
 \label{eq:m1-forward}
\end{equation}
where $\eta_m$ is the LoRA scale and $q_i^v$ weights task $i$'s
visual adapter. Each projector transformation has its own factors.
Training activates the current adapter, $q^v=e_t$, where $e_t$ is
the $t$th one-hot basis vector. During inference,
image-feature similarity to retained task features determines $q^v$.

For language layer $\ell$, the adapter combines a shared input
factor $A_{\ell s}$ with task-wise factors $(A_{\ell i},B_{\ell i})$,
\begin{equation}
 \begin{aligned}
 \Delta_{\ell i}(h)&=B_{\ell i}(A_{\ell s}+\lambda A_{\ell i})h,\\
 y_\ell&=W_{\ell0}h+\eta\sum_{i=1}^{t}r_{\ell i}\Delta_{\ell i}(h).
 \end{aligned}
 \label{eq:expert}
\end{equation}
Here, $W_{\ell0}$ is the frozen pretrained weight,
$A_{\ell s},A_{\ell i}\in\mathbb R^{r\times d_{\rm in}}$, and
$B_{\ell i}\in\mathbb R^{d_{\rm out}\times r}$. Here, $r$ is the
adapter rank, $d_{\rm in}$ and $d_{\rm out}$ are input and output
dimensions, and $h\in\mathbb R^{d_{\rm in}}$ is the layer input.
The shared factor captures transformations common to multiple tasks,
and the task-specific factors encode specialized adaptations.
The coefficients $\eta$ and $\lambda$ control their scale, while
$r_{\ell i}$ determines the contribution of adapter $i$ to the input.
Further details about adapter settings and task-feature construction are provided in Appendix~\ref*{app-main:c_protocol}.

\subsection{History-Guided Mapping Adaptation}
\label{sec:hma}
The projector interventions show that historical mappings remain
useful for earlier tasks. HMA uses these mappings to guide the
learning of new projector adapters. For transformation $j$, we
construct references $X^m_{j,A,<t}$ and $X^m_{j,B,<t}$ from the
completed projector factors. Following Conceptor
theory~\citep{jaeger2017conceptors} and CODE-CL~\citep{codecl2025},
a reference $X\in\mathbb R^{n\times d}$ defines the soft constraint
\begin{equation}
 \begin{aligned}
 C_X&=X^\top[XX^\top+(na^{-2}+\epsilon)I_n]^{-1}X,\\
 \mathcal F(G;X)&=G(I_d-\rho C_X),
 \end{aligned}
 \label{eq:soft-protection}
\end{equation}
where $X$ contains $n$ reference vectors of dimension $d$, $G$ is a
gradient matrix with $d$ columns, and $I_k$ is the $k$-dimensional
identity. The aperture $a>0$ and strength $\rho\in[0,1]$ control the
constraint, while $\epsilon>0$ stabilizes the inverse.
The operator $C_X$ assigns larger weights to directions strongly
represented in $X$. Applying $\mathcal F$ therefore attenuates gradient
components along those directions.

HMA applies this constraint to the current projector factors
$P\in\{A,B\}$,
\begin{equation}
 \widetilde{\nabla P^m_{j,t}}
 =\mathcal F(\nabla P^m_{j,t};X^m_{j,P,<t}).
 \label{eq:m1}
\end{equation}
The two factors use references in their respective input and low-rank
coordinates. This reduces updates along historical mapping directions
while allowing the new adapter to learn the incoming task. Further details about historical reference construction are provided in Appendix~\ref*{app-main:c_protocol}.

\subsection{Selection-Guided Adapter Reuse}
\label{sec:sar}
The expert interventions show that retained language adapters can
continue to support earlier tasks. SAR incorporates related historical
adapters into new-task learning and maintains prompt-conditioned
selection during generation.

Let $h^q_\ell$ denote the current prompt feature and $\mu_{\ell i}$
the representative prompt feature for task $i$. During training,
we select the most related historical adapter,
$j^\star_{<t}=\operatorname*{arg\,max}_{i<t}\cos(h^q_\ell,\mu_{\ell i})$, and combine
its output with the current adapter,
\begin{equation}
 \Delta^{\rm train}_\ell(h)=
 \beta\Delta_{\ell t}(h)+(1-\beta)
 \operatorname{sg}[\Delta_{\ell j^\star_{<t}}(h)].
 \label{eq:coactivation}
\end{equation}
Here, $\beta$ controls the contribution of the current adapter, and
$\operatorname{sg}$ stops gradient flow through the historical
contribution. The resulting prediction trains the current adaptation
in the presence of related historical knowledge. The first task uses
its current adapter alone.

During inference, SAR selects a language adapter from the prompt
feature at each adapted layer and retains that selection throughout
the answer,
\begin{equation}
 \begin{aligned}
 i^\star_\ell&=\operatorname*{arg\,max}_{i\leq t}\cos(h^q_\ell,\mu_{\ell i}),\\
 \boldsymbol r_{\ell,u}&=e_{i^\star_\ell}\quad (u=1,\ldots,U),
 \end{aligned}
 \label{eq:m2}
\end{equation}
where $U$ is the response length and $\boldsymbol r_{\ell,u}$ collects
the adapter weights $r_{\ell i}$ at token $u$. The one-hot vector $e_i$
selects adapter $i$. Layer-wise selection allows different layers to use different
retained adaptations, while keeping each layer's selection consistent
across response tokens. Further details about feature summaries and adapter selection are provided in Appendix~\ref*{app-main:c_protocol}.

\subsection{Pretrained Subspace Guidance}
\label{sec:psg}
PSG guides language adaptation with directions from the pretrained
model. At layer $\ell$, it constructs $X^0_\ell$ from rows of the
frozen weight $W_{\ell0}$ and applies the soft constraint to the
shared and current task-wise input factors,
\begin{equation}
 \begin{aligned}
 \widetilde{\nabla A_{\ell s}}&=\mathcal F(\nabla A_{\ell s};X^0_\ell),\\
 \widetilde{\nabla A_{\ell t}}&=\mathcal F(\nabla A_{\ell t};X^0_\ell).
 \end{aligned}
 \label{eq:m3}
\end{equation}
Both factors use the same input coordinates as the pretrained
reference. HMA and PSG thus share a gradient operator, with historical
projector factors guiding visual adaptation and pretrained language
weights guiding language adaptation.

We optimize the autoregressive instruction loss and apply the HMA
and PSG constraints before each parameter update. Further results about the spectral properties of the gradient constraint are provided in Appendix~\ref*{app-main:f_analysis}.

\begin{algorithm}[t]
\caption{\method}\label{alg:mapselect}
\small
\algrenewcommand\algorithmicrequire{\textbf{Input}}
\algrenewcommand\algorithmicensure{\textbf{Output}}
\algrenewcommand\algorithmicindent{1em}
\begin{algorithmic}[1]
\Require Tasks $\{\mathcal D_t\}_{t=1}^{T}$, pretrained MLLM.
\Ensure Task-wise adapters and feature summaries.
\State Initialize shared language factors and task summaries.
\For{$t=1,\ldots,T$}
 \State Initialize task-$t$ projector and language adapters.
 \State Construct historical and pretrained references.
 \For{each minibatch in $\mathcal D_t$}
  \State Activate the current projector adapter.
  \State Select a related old language adapter, if available.
  \State Compute the instruction loss using SAR, Eq.~\eqref{eq:coactivation}.
  \State Compute gradients of the trainable adapter factors.
  \State Apply HMA to projector gradients, Eq.~\eqref{eq:m1}.
  \State Apply PSG to language gradients, Eq.~\eqref{eq:m3}.
  \State Update trainable factors and task feature summaries.
 \EndFor
 \State Retain task-$t$ adapters and summaries.
\EndFor
\end{algorithmic}
\smallskip
\textbf{Inference.} Select visual adapters from image features and language
adapters from prompt features. Keep language selection fixed throughout
the answer, Eq.~\eqref{eq:m2}.
\end{algorithm}

\section{Experiments}
\label{sec:experiments}
\definecolor{mapresultshade}{RGB}{229,246,252}

\subsection{Experimental Setup}
\noindent\textbf{Benchmarks.}
We evaluate \method{} on Unseen Continual Instruction Tuning
(UCIT)~\citep{hide2025,mcitlib2025} and MLLM Domain-Continual
Learning (MLLM-DCL)~\citep{mllmcl2025}. UCIT follows ImageNet-R (IN-R), ArxivQA
(AQA), VizWiz (VW), IconQA (IQ), CLEVR-Math (CM), and Flickr30k
(F30K). MLLM-DCL follows Remote Sensing (RS), Medicine (Med),
Autonomous Driving (AD), Science (Sci), and Finance (Fin).

\medskip
\noindent\textbf{Baselines.}
We include LoRA-FT~\citep{hu2022lora}, O-LoRA~\citep{olora2023},
MoELoRA~\citep{coin2024}, Replay~\citep{rolnick2019replay},
ModalPrompt~\citep{modalprompt2025}, CL-MoE~\citep{clmoe2025},
HiDe~\citep{hide2025}, SEFE~\citep{sefe2025},
DISCO~\citep{fcit2025}, SMoLoRA~\citep{smolora2025},
SAME~\citep{same2026}, DGG~\citep{dgg2025}, and
CRAM~\citep{tang2026cram} in both benchmark comparisons.

\medskip
\noindent\textbf{Implementation and metrics.}
We use LLaVA-v1.5-7B~\citep{llava15_2024} as the backbone and
adapt its projector and language model with LoRA. Following
MCITlib~\citep{mcitlib2025}, we evaluate continual learning using
Mean Finetune Accuracy (MFT), Mean Final Accuracy (MFN), Mean
Average Accuracy (MAA), and Backward Transfer (BWT). MFT measures
accuracy immediately after each task is learned. Task scores follow
the benchmark protocols, including CaptionAvg for captioning.
Results in Tables~\ref{tab:ucit} and~\ref{tab:dcl} are averaged over
five random seeds,
$\{37,42,137,1337,13337\}$. Further details about training settings and metric definitions,
and further results about ArxivQA prompt sensitivity, are provided in
Appendix~\ref*{app-main:c_protocol}.

\subsection{Main Results}
\begin{table*}[t]
\captionsetup{labelfont={bf,color=black},textfont={color=black},justification=raggedright,singlelinecheck=false,skip=4pt}
\centering
\caption{UCIT results after six tasks. QA uses accuracy (\%), and VW/F30K use CaptionAvg. Bold/underline indicate the best/second-best measured results.}
\label{tab:ucit}
\fontsize{8.5}{10}\selectfont
\renewcommand{\arraystretch}{1.12}
\setlength{\tabcolsep}{3.0pt}
\begin{tabularx}{\textwidth}{@{}l*{10}{>{\raggedleft\arraybackslash}X}@{}}
\toprule
Method & IN-R & AQA & VW & IQ & CM & F30K & MFT ($\uparrow$) & MFN ($\uparrow$) & MAA ($\uparrow$) & BWT ($\uparrow$)\\
\midrule
LoRA-FT & 64.87 & 77.30 & 44.75 & 70.03 & 55.87 & 57.24 & \underline{76.44} & 61.68 & 76.96 & -17.72\\
O-LoRA & 85.27 & 93.97 & 37.43 & 66.73 & 53.30 & 52.09 & 70.80 & 64.80 & 78.82 & -7.21\\
MoELoRA & 64.43 & 78.50 & 44.50 & 64.17 & 56.20 & 57.39 & 75.30 & 60.86 & 76.76 & -17.32\\
Replay & 75.63 & 84.50 & 50.86 & 71.03 & 67.10 & \underline{57.53} & 75.60 & 67.78 & 79.62 & -9.39\\
ModalPrompt\, & 74.13 & 82.21 & 44.20 & 63.03 & 52.92 & 54.31 & 70.15 & 61.80 & 71.09 & -10.20\\
CL-MoE & 65.57 & 77.63 & 44.77 & \underline{71.87} & 47.83 & 57.52 & 75.89 & 60.87 & 76.73 & -18.03\\
HiDe & 85.37 & 91.80 & 47.44 & 67.47 & 55.23 & 53.96 & 72.79 & 66.88 & 79.78 & -7.10\\
SEFE\, & 84.13 & 91.19 & 50.05 & 68.98 & 62.07 & 55.56 & 74.23 & 68.66 & 78.11 & -6.80\\
DISCO & \underline{87.70} & \underline{94.50} & 46.69 & 70.93 & 55.67 & 56.54 & 76.24 & 68.67 & \underline{81.82} & -9.08\\
SMoLoRA & 75.53 & 73.50 & 45.63 & 71.10 & 60.37 & 56.24 & 72.36 & 63.73 & 75.10 & -10.36\\
SAME\, & 87.20 & 94.27 & \underline{59.32} & 67.88 & \underline{70.06} & 56.04 & 75.05 & \underline{72.46} & 81.23 & \underline{-3.40}\\
DGG\, & 86.17 & 93.24 & 55.44 & 69.82 & 65.19 & 56.02 & 75.37 & 70.98 & 80.36 & -5.20\\
CRAM\, & 78.32 & 87.18 & 48.51 & 61.10 & 54.18 & 52.13 & 70.80 & 63.57 & 73.38 & -8.10\\
\rowcolor{mapresultshade}
\textbf{\method{}} & \textbf{89.07} & \textbf{94.90} & \textbf{64.82} & \textbf{75.37} & \textbf{74.53} & \textbf{59.11} & \textbf{77.15} & \textbf{74.52} & \textbf{83.26} & \textbf{0.06}\\
\bottomrule
\end{tabularx}
\end{table*}

\begin{table*}[t]
\captionsetup{labelfont={bf,color=black},textfont={color=black},justification=raggedright,singlelinecheck=false,skip=4pt}
\centering
\caption{MLLM-DCL results after five domains. Scores use the domain-specific benchmark metrics. Bold/underline indicate the best/second-best measured results.}
\label{tab:dcl}
\fontsize{8.5}{10}\selectfont
\renewcommand{\arraystretch}{1.12}
\setlength{\tabcolsep}{3.0pt}
\begin{tabularx}{\textwidth}{@{}l*{9}{>{\raggedleft\arraybackslash}X}@{}}
\toprule
Method & RS & Med & AD & Sci & Fin & MFT ($\uparrow$) & MFN ($\uparrow$) & MAA ($\uparrow$) & BWT ($\uparrow$)\\
\midrule
LoRA-FT & 67.86 & 47.53 & 34.24 & 41.88 & \underline{89.89} & 66.52 & 56.28 & 62.24 & -12.80\\
O-LoRA & 75.78 & 36.69 & 23.05 & 37.05 & 81.02 & 59.33 & 50.72 & 59.45 & -10.77\\
MoELoRA & 77.41 & 46.57 & 31.02 & 40.58 & 88.23 & 65.65 & 56.76 & 62.58 & -11.12\\
Replay\, & 74.32 & 50.09 & 39.13 & 44.32 & 89.70 & 66.34 & 59.51 & 64.21 & -8.50\\
ModalPrompt & 48.65 & 28.35 & 12.06 & 30.12 & 82.82 & 54.55 & 40.40 & 42.19 & -17.69\\
CL-MoE & 71.41 & 48.56 & 30.56 & 41.74 & 89.79 & 66.60 & 56.41 & 62.39 & -12.74\\
HiDe & 71.63 & 48.73 & 35.06 & 40.86 & 78.06 & 60.75 & 54.87 & 61.12 & -7.35\\
SEFE & 77.28 & 51.97 & 41.63 & 42.08 & 89.36 & 66.38 & 60.46 & 65.14 & -7.40\\
DISCO & 76.67 & 44.85 & 47.01 & 49.07 & 89.07 & 64.93 & 61.34 & 64.29 & -4.50\\
SMoLoRA\, & 73.67 & 48.14 & 36.36 & 43.13 & 88.19 & 65.13 & 57.90 & 63.32 & -9.25\\
SAME\, & \underline{78.26} & \underline{54.17} & \underline{49.14} & \underline{50.32} & 89.68 & \underline{67.15} & \underline{64.31} & \underline{67.87} & \underline{-3.75}\\
DGG\, & 77.24 & 53.32 & 47.25 & 49.45 & 89.55 & 66.46 & 63.36 & 66.65 & \underline{-3.75}\\
CRAM\, & 70.67 & 45.18 & 33.15 & 40.35 & 86.68 & 62.57 & 55.21 & 60.30 & -9.01\\
\rowcolor{mapresultshade}
\textbf{\method{}} & \textbf{79.59} & \textbf{56.77} & \textbf{52.49} & \textbf{52.42} & \textbf{90.13} & \textbf{67.39} & \textbf{66.28} & \textbf{68.28} & \textbf{-1.39}\\
\bottomrule
\end{tabularx}
\end{table*}

\textbf{\method{} improves final performance across both task and
domain sequences, as shown in Tables~\ref{tab:ucit} and~\ref{tab:dcl}.}
As illustrated in Figure~\ref{fig:benchmark-radar}, \method{} achieves
higher MFN and BWT than all 13 baselines on both benchmarks. On UCIT,
it reaches 74.52 MFN compared with 72.46 for SAME.
The gains are particularly evident on VizWiz and CLEVR-Math, where
\method{} reaches 64.82 and 74.53, respectively. These tasks require
different forms of visual reasoning and generation, indicating that
the improvement extends beyond a single output format. ImageNet-R
and ArxivQA also benefit, with final accuracies of 89.07\% and
94.90\%. Together, these results support retaining visual adaptations
that remain useful as subsequent tasks change the input distribution.

The learning and retention metrics complement the final task scores.
\method{} achieves 83.26 MAA and 0.06 BWT on UCIT, compared with
$-3.40$ BWT for SAME. MAA measures performance throughout the
sequence, while BWT compares earlier-task scores before and after
subsequent learning.

The improvement extends to the five specialized domains in MLLM-DCL.
\method{} achieves 66.28 MFN and $-1.39$ BWT, compared with
64.31 MFN and $-3.75$ BWT for SAME. It also obtains higher final scores
in every domain. Its MAA of 68.28 further shows
that the benefit extends across the learning sequence. The largest
gains over DISCO occur in medicine and autonomous driving, where
changes in visual content accompany specialized instructions. These
results are consistent with treating the visual mapping and language
adaptations as coupled parts of continual multimodal learning.
Further results about stage-wise benchmark performance are provided
in Appendix~\ref*{app-main:d_benchmarks}.

\subsection{Ablation Study}
\label{sec:component-analysis}
\begin{table}[t]
\captionsetup{labelfont={bf,color=black},textfont={color=black},justification=raggedright,singlelinecheck=false,skip=4pt}
\centering
\small
\renewcommand{\arraystretch}{1.15}
\setlength{\tabcolsep}{2pt}
\caption{Component ablation on UCIT. Best results are bold.}
\label{tab:ablation}
\begin{tabularx}{\columnwidth}{@{}*{3}{>{\centering\arraybackslash}p{23pt}}*{4}{>{\raggedleft\arraybackslash}X}@{}}
\toprule
HMA & SAR & PSG & MFT ($\uparrow$) & MFN ($\uparrow$) & MAA ($\uparrow$) & BWT ($\uparrow$)\\
\midrule
-- & $\checkmark$ & $\checkmark$ & {74.96} & 73.97 & 77.60 & -1.91\\
$\checkmark$ & -- & $\checkmark$ & 75.08 & 73.80 & {82.15} & -1.89\\
$\checkmark$ & $\checkmark$ & -- & 75.14 & 73.84 & 80.53 & -1.55\\
\rowcolor{mapresultshade}
$\checkmark$ & $\checkmark$ & $\checkmark$ & \textbf{75.57} & \textbf{74.13} & \textbf{82.86} & \textbf{-1.13}\\
\bottomrule
\end{tabularx}
\end{table}

\textbf{The complete model achieves the strongest final performance
and backward transfer in the component ablation.}
As shown in Table~\ref{tab:ablation}, removing HMA, SAR, or PSG
lowers MFN, while removing each component also reduces BWT.
The components act on different parts of the learning process:
HMA guides visual adaptation using historical mappings, SAR brings
historical language knowledge into current-task learning, and PSG
constrains language updates using pretrained directions. Their
combination therefore addresses both the representations supplied
to retained adapters and the adaptations learned from those representations.

As a capacity control, Base-Fixed retains the same adapter
banks and inference selection but removes HMA, historical training
reuse, and PSG. The full model improves final MFN by 0.69 points
in the paired comparison, supporting a contribution from training
beyond the retained parameter budget. Further results about
capacity-matched comparisons are provided in Appendix~\ref*{app-main:e_controlled}.

\subsection{Mapping and Adapter Analysis}
\begin{table}[t]
\captionsetup{labelfont={bf,color=black},textfont={color=black},justification=raggedright,singlelinecheck=false,skip=4pt}
\centering
\small
\renewcommand{\arraystretch}{1.15}
\setlength{\tabcolsep}{3pt}
\caption{Frozen and trainable projectors on UCIT. Bold indicates the better score within each learner.}
\label{tab:frozen-projector}
\begin{tabularx}{\columnwidth}{@{}ll*{2}{>{\raggedleft\arraybackslash}X}@{}}
\toprule
Learner & Projector & MFT ($\uparrow$) & MFN ($\uparrow$)\\
\midrule
LoRA-FT & Trainable & \textbf{76.34} & \textbf{61.09}\\
 & Frozen & 74.69 & 61.08\\
\midrule
O-LoRA & Trainable & \textbf{71.77} & \textbf{64.90}\\
 & Frozen & 59.10 & 55.90\\
\midrule
\rowcolor{mapresultshade}
\textbf{\method{}} & Trainable & \textbf{75.46} & \textbf{73.85}\\
\rowcolor{mapresultshade}
 & Frozen & 74.81 & 72.47\\
\bottomrule
\end{tabularx}
\end{table}

\begin{figure*}[t]
\centering
\includegraphics[width=\textwidth]{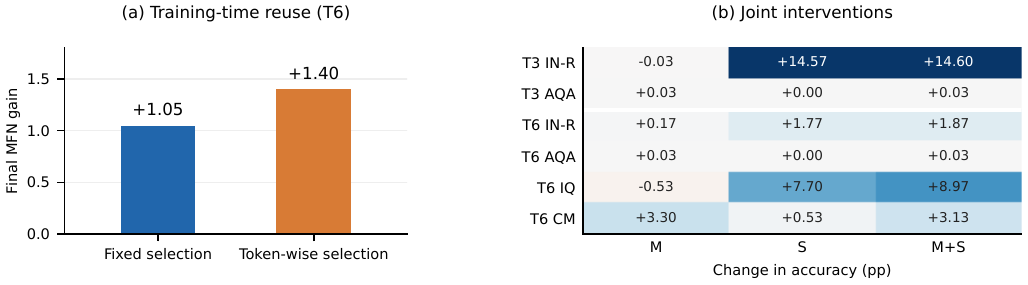}
\caption{\textbf{Historical reuse and joint mapping--selection interventions.}
(a) The change in mean task score from enabling historical reuse,
with fixed or token-wise inference selection, at the final stage T6.
(b) Changes from native predictions under mapping ($M$), language-adapter
selection ($S$), and their combination ($M+S$). Mapping-only
interventions keep native language routes. Each panel retains its
own controlled configuration. Forced interventions measure retained
capability under the specified choices.}
\label{fig:mechanism-analysis}
\end{figure*}

\begin{figure*}[t]
\centering
\includegraphics[width=\textwidth]{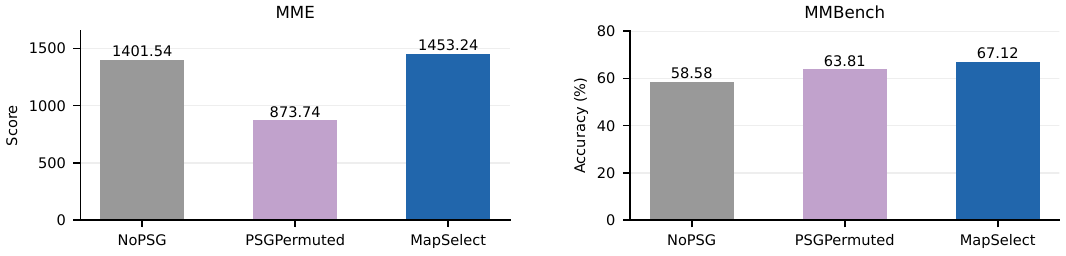}
\caption{\textbf{Retained pretrained capability after the UCIT sequence.}
MME and MMBench scores at T6, using the same candidate-scoring
protocol for all variants. NoPSG removes pretrained guidance,
and PSGPermuted changes the reference directions.}
\label{fig:pretrained-analysis}
\end{figure*}

\noindent\textbf{Visual adaptation.}
Keeping the projector frozen is a simple way to prevent changes in
the visual mapping, but it also removes the ability to adapt that
mapping to incoming tasks. As shown in
Table~\ref{tab:frozen-projector}, trainable projectors improve
O-LoRA's final MFN by 9.00 points and \method{}'s by 1.38 points.
LoRA-FT has almost identical final means under the two settings.
For \method{}, the comparison supports learning new
visual adaptations while retaining historical alternatives.

This distinction connects the training comparison to the projector
interventions in Figure~\ref{fig:mapping}. A historical projector
can remain useful for an earlier task even when further adaptation
is beneficial for the incoming task. Freezing one mapping for the
entire sequence and retaining multiple learned mappings address
different objectives. In a shared-projector control, \method{}
achieves 73.90 on CLEVR-Math at T6, compared with 43.00 when all
tasks share one projector. This result illustrates why maintaining
separate visual adaptations can be useful when tasks impose different
requirements on the image-to-language interface. Further results
about visual mapping adaptation are provided in Appendix~\ref*{app-main:f_controls}.

\medskip
\noindent\textbf{Historical adapter reuse.}
We separate the effect of reusing historical adapters during training
from the choice of inference-time selection. As illustrated in
Figure~\ref{fig:mechanism-analysis}(a), reuse improves final MFN
under both fixed and token-wise selection. The improvement under
two inference rules supports learning with historical adaptations.
This comparison evaluates performance after the full task sequence.

Fixed selection preserves each layer's adapter identity throughout
an answer. Token-wise selection gives a slightly higher final score
in this comparison. Further results about
historical reuse and inference selection are provided in Appendix~\ref*{app-main:f_controls}.

\medskip
\noindent\textbf{Joint mapping and selection.}
We next intervene on both parts of the model while keeping its
learned parameters fixed. As illustrated in
Figure~\ref{fig:mechanism-analysis}(b), the relative contributions
depend on the task. On ImageNet-R at T3, the gain primarily comes
from selecting the retained language adapter. On CLEVR-Math at T6,
the visual mapping contributes the larger improvement. IconQA
benefits from their combination: selecting its language adapter
increases accuracy from 66.20\% to 73.90\%, and using its visual
mapping as well increases accuracy to 75.17\%.

These comparisons connect visual representations to the language
adaptations that process them. Mapping-only interventions replay the
native language routes, whereas joint interventions change both factors.
On IconQA, their interaction is positive: the combined gain exceeds the
sum of individual gains by 1.80 percentage points. The relative
contributions vary across tasks, supporting complementary roles for
mapping preservation and adapter selection.
Further results about joint interventions are provided in Appendix~\ref*{app-main:f_controls}.

\subsection{Pretrained Knowledge Preservation}

We evaluate PSG on MME~\citep{mme2023} and MMBench~\citep{mmbench2023}
after UCIT to examine retained capabilities beyond the continual-training
tasks. All variants use the same candidate-scoring protocol. NoPSG
removes the constraint, while PSGPermuted tests its reference directions.

As illustrated in Figure~\ref{fig:pretrained-analysis}, \method{}
outperforms both variants. Relative to NoPSG, it improves MMBench by
8.54 percentage points and MME by 51.69 points. The lower scores with
permuted references support using pretrained weight directions to guide
adaptation, consistent with the component ablation.
Further results about pretrained retention are provided in Appendix~\ref*{app-main:g_scope_cost}.

\subsection{Computational Cost}
The UCIT model retains 1.55B adapter parameters after six tasks,
requiring 3.10 GB in bf16. This includes 240.07M task-specific
parameters per task and 109.31M shared parameters. The backbone
is shared, while adapter storage grows with the sequence. During
generation, each language layer uses one selected task adapter.

Training the full model takes 4.49 hours on eight A800 GPUs,
compared with 2.58 hours for Base-Fixed. Historical reuse evaluates
an earlier adaptation during learning, and the gradient constraints
add computation before the parameter update.
Further results about inference latency, parameter budgets, and
baseline costs are provided in Appendix~\ref*{app-main:g_scope_cost}.

\FloatBarrier
\needspace{7\baselineskip}
\section{Conclusion}
\label{sec:conclusion}
We propose \method{} for continual multimodal instruction tuning.
Projector and expert interventions identify mapping preservation and
adapter selection as complementary factors in retaining earlier-task
performance. HMA uses historical projector directions to guide visual
adaptation, SAR incorporates related language adapters into learning
and generation, and PSG constrains updates with pretrained knowledge.
Experiments on UCIT and MLLM-DCL demonstrate improved final performance
and retention. Ablations and controlled comparisons further establish
the roles of visual adaptation, historical reuse, and pretrained
guidance.

\FloatBarrier
\clearpage
\section*{LLM Usage Statement}
OpenAI ChatGPT and Codex assisted with literature searches, code
development and debugging, experiment-workflow scripting, figure and
table preparation, and manuscript editing and LaTeX formatting.
The authors reviewed the AI-assisted outputs and take full
responsibility for the results, citations, and claims.
\bibliographystyle{ieeenat_fullname}
\bibliography{references}
\end{document}